\documentclass[10pt,twocolumn,letterpaper]{article}

\usepackage[pagenumbers]{iccv}      

\usepackage{graphicx}	
\usepackage{amsmath}	
\usepackage{amssymb}	
\usepackage{booktabs}
\usepackage{times}
\usepackage{microtype}
\usepackage{epsfig}
\usepackage[table,xcdraw,dvipsnames]{xcolor}
\usepackage{caption}
\usepackage{float}
\usepackage{placeins}
\usepackage{color, colortbl}
\usepackage{stfloats}
\usepackage{enumitem}
\usepackage{tabularx}
\usepackage{xstring}
\usepackage{multirow}
\usepackage{xspace}
\usepackage{url}
\usepackage{subcaption}
\usepackage[hang,flushmargin]{footmisc}
\usepackage{algorithm,algorithmic}
\usepackage{float}
\usepackage{pifont}
\newcommand{\partitle}[1]{\smallskip \noindent \textbf{#1.}}

\definecolor{darkgreen}{rgb}{0.0, 0.5, 0.0} 
\definecolor{darkred}{rgb}{0.5, 0.0, 0.0}

\newcommand{\projectname}{\textsc{SHAPE}}

\newcommand{\todo}[1]{{\color{red}#1}}

\definecolor{iccvblue}{rgb}{0.21,0.49,0.74}
\usepackage[pagebackref,breaklinks,colorlinks,allcolors=iccvblue]{hyperref}

\def\paperID{7530} 
\def\confName{ICCV}
\def\confYear{2025}

\title{\projectname~ : Self-Improved Visual Preference Alignment by Iteratively Generating Holistic Winner}

\author{Kejia Chen\thanks{Equal contribution.}\textsuperscript{\rm ~~1}~~~Jiawen Zhang\footnotemark[1]\textsuperscript{\rm ~~1}~~~Jiacong Hu\textsuperscript{\rm 1}~~~Jiazhen Yang\textsuperscript{\rm 1~~~}  Jian Lou\textsuperscript{\rm 2}~~~Zunlei Feng\textsuperscript{\rm 1}~~~Mingli Song\textsuperscript{\rm 1} \\\\
\textsuperscript{\rm 1} Zhejiang Univerisity 
\hspace{0.3cm} \textsuperscript{\rm 2}Sun Yat-sen University 
}

\begin{document}
\maketitle
\begin{abstract}
Large Visual Language Models (LVLMs) increasingly rely on preference alignment to ensure reliability, which steers the model behavior via preference fine-tuning on preference data structured as ``image - winner text - loser text'' triplets. However, existing approaches often suffer from limited diversity and high costs associated with human-annotated preference data, hindering LVLMs from fully achieving their intended alignment capabilities. We present \projectname, a self-supervised framework capable of transforming the already abundant supervised text-image pairs into holistic preference triplets for more effective and cheaper LVLM alignment, eliminating the need for human preference annotations. 
Our approach facilitates LVLMs in progressively enhancing alignment capabilities through iterative self-improvement. The key design rationale is to devise preference triplets where the winner text consistently improves in holisticness and outperforms the loser response in quality, thereby pushing the model to ``strive to the utmost'' of alignment performance through preference fine-tuning. For each given text-image pair, \projectname~ introduces multiple visual augmentations and pairs them with a summarized text to serve as the winner response, while designating the original text as the loser response. Experiments across \textbf{12} benchmarks on various model architectures and sizes, including LLaVA and DeepSeek-VL, show that \projectname~ achieves significant gains, for example, achieving +11.3\% on MMVet (comprehensive evaluation), +1.4\% on MMBench (general VQA), and +8.0\% on POPE (hallucination robustness) over baselines in 7B models. Notably, qualitative analyses confirm enhanced attention to visual details and better alignment with human preferences for holistic descriptions.

\end{abstract}    
\section{Introduction}
\label{sec:intro}

Inheriting the breakthroughs in large language models (LLMs), large visual language models (LVLMs) have developed rapidly and achieved astounding performance on a wide range of image-text understanding tasks, showcasing exceptional potential \citep{alayrac2022flamingo, zhu2023minigpt,liu2024improved,liu2024visual}.
Despite their remarkable capabilities, there persists significant challenges in aligning LVLMs' generations with human preferences, where models produce texts that diverge from visual inputs and human expectations \cite{sun2023aligning,chen2023sharegpt4v}. Failure to achieve such preference alignments could lead to serious consequences for LVLMs' real-world applications, ranging from the spread of misinformation to the risk of harmful decision-making.

Given the critical role of preference alignment, there has been a surge in research interest in addressing the hurdles of aligning LVLMs' generations with human preferences.
The seminal approach involves first collecting large-scale preference data and human annotations, then utilizing preference fine-tuning algorithms such as reinforcement learning with human feedback (RLHF) or more recent direct preference optimization (DPO) based on preference triplets \citep{banerjee2020weaqa,changpinyo2022all, sun2023aligning,li2025llama}. This approach can be resource-intensive, as human annotation can be demanding and costly. Additionally, it may struggle to achieve comprehensive coverage of all possible distributions of human preferences.
To remedy these limitations in the requirement of human-annotated preference data, alternative approaches propose leveraging external LVLMs like GPT-4 to generate preference data. However, this approach introduces new challenges: inconsistent scoring mechanisms and discrete metrics hinder accurate capability assessment \citep{bai2022training, jiang2023llm, yuan2024self}. Additionally, relying on external models for scoring is both costly and limited in capturing the nuanced semantic relationships between visual and textual content. As a result, the preference alignment efforts of existing LVLMs are hindered by the challenge of acquiring high-quality and cost-efficient preference data.


Recognizing the scarcity of preference data as a critical bottleneck \citep{wu2024self, xiong2024iterative} limiting LVLMs alignment performance, recent studies have sought to mitigate this challenge, primarily by generating LVLMs-aligned preference data using LLMs to reduce reliance on labeled data. For instance, SIMA \cite{wang2024enhancing} extracts images and question from visual input and lets the model generate responses, and then uses a self-assessment scheme to create preference pairs for training. While CSR \cite{zhou2024calibrated} implements a reward mechanism for model self-generation and calibration. SEVA \cite{zhu2024self} incorporates DPO for preference alignment but relies solely on comparing raw and enhanced versions of the model's visual outputs, limiting its ability to capture rich semantic connections.

By and large, these recent attempts devise the preference data by exclusively focusing on introducing multiple text-side operations, while leaving the image-side fixed during the winner and loser samples generation \cite{lee2023rlaif}, which overlooks the crucial visual-textual semantic interplay. Our experiments demonstrate that LVLMs are highly sensitive to image-side modifications, such as random flips, which can lead to substantial semantic shifts in model outputs \citep{he2020momentum,zhu2023multi,zhu2023coarse,zhu2024self}. 
This \hyperref[fig:att]{observation} transforms a perceived limitation into an opportunity. This sensitivity, rather than being a drawback, offers a supervisory signal to generate diverse, semantically related text responses without additional human preference annotations, thereby yielding high-quality and cost-efficient preference triplets for LVLMs' alignment. By leveraging visual augmentation-induced variations to create preference pairs automatically, we can construct rich semantic comparisons while preserving the crucial visual-textual semantic interplay.
Therefore, we address a fundamental question in LVLMs preference alignment: \textit{How can we leverage existing supervised annotation data to automatically construct high-quality preference pairs while maintaining low annotation costs and ensuring holistic semantic understanding?}

\begin{figure}[tp]
    \centering
    \includegraphics[width=\linewidth]{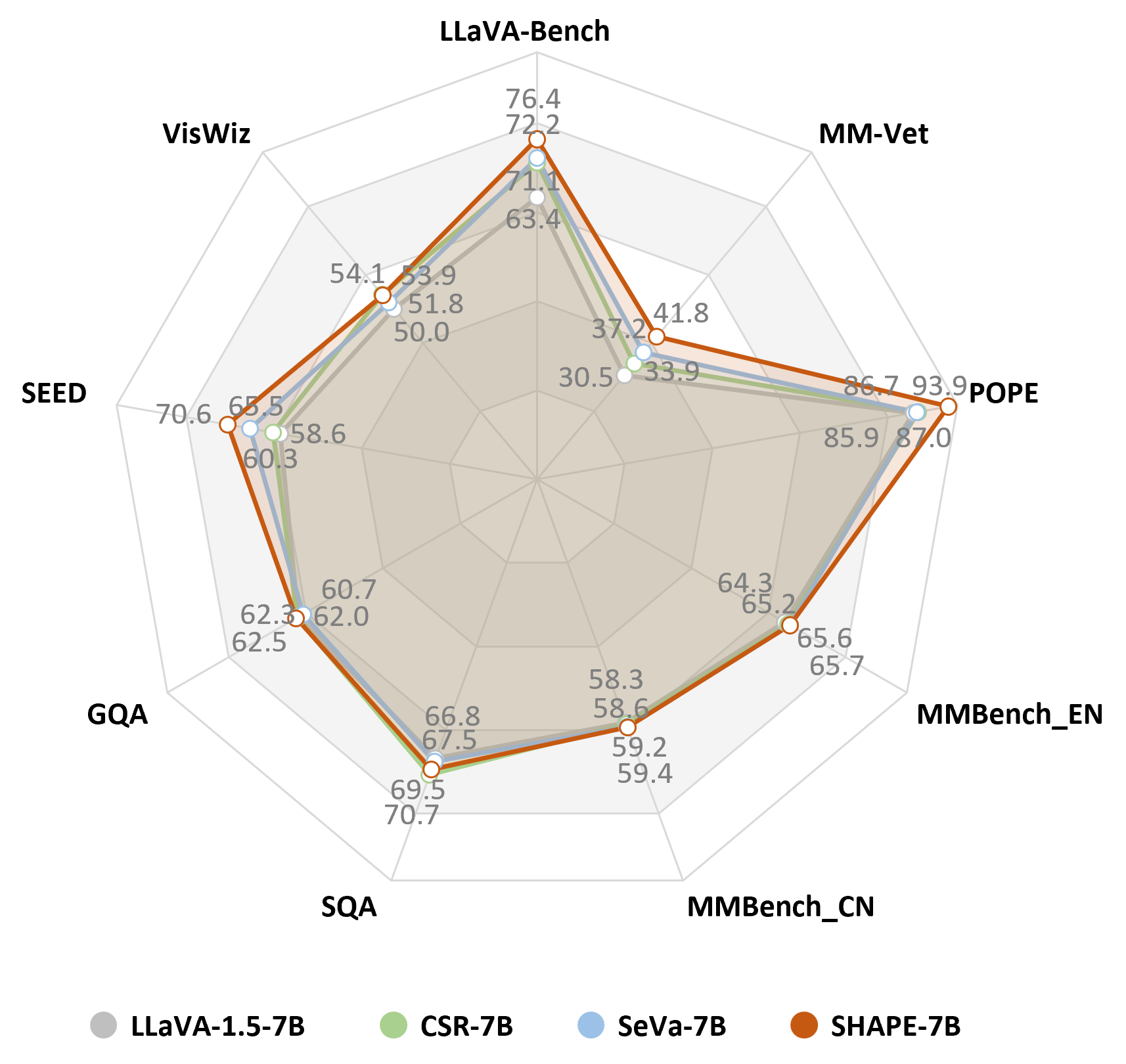}
    \caption{Comprehensive evaluation of \projectname~'s enhanced performance against SOTA models across multimodal benchmarks: Achieving +11.3\% on MMVet, +1.4\% on MMBench, and +8.0\% on POPE over baselines in 7B Models.}
    \label{fig:benchmark}
\end{figure}

To tackle these challenges, we propose \textbf{\projectname~} (\textbf{\underline{S}}elf-improved \textbf{\underline{H}}olistic \textbf{\underline{A}}lignment for \textbf{\underline{P}}reference \textbf{\underline{E}}nhancement), a novel framework that optimizes LVLMs preference alignment through automated construction of competitive preference pairs. Our design rationale stems from a key observation: effective preference alignment requires carefully crafted pairs that serve two critical properties: \textit{\textbf{1. A winner text that exemplifies superior quality; 2. A competitive loser that is already capable of providing meaningful enough learning signals.}} This deliberate construction ensures effective preference alignment while eliminating the need for costly manual annotations.

\noindent 
\partitle{Main Contributions and Results} 
Recognizing the inherent limitations in the knowledge depth and reasoning capabilities of single LVLMs, as well as the negative responses triggered by image-side augmentation strategies, which highlight the need for a more robust approach to preference alignment, we propose \projectname~, a novel self-improved framework that leverages the intrinsic relationships between visual and textual modalities. By synthesizing stronger responses through summarization and augmentation-aware mechanisms, \projectname~ ensures both robustness and semantic coherence. Furthermore, our framework constructs holistic preference pairs by generating ``winners" from diverse candidate outputs, dynamically improving alignment with preferred behaviors while minimizing annotation costs. 
Extensive evaluations conducted across 12 benchmarks demonstrate the effectiveness of our approach, achieving substantial performance gains of of 10.36\%, 6.46\%, 2.3\% and 3.3\% on the MMVet benchmark for LLaVA-1.5-7B, 13B and DeepSeek-VL2-3B, 27B model variants respectively, thereby surpassing state-of-the-art results while effectively addressing the limitations of single-model reasoning and the challenges associated with augmentation strategies.
\section{Related Work}
\label{sec:related}

\subsection{Large Visual Language Models}

Recent years have witnessed remarkable progress in Visual Language Models (LVLMs), from early pioneers like CLIP to advanced models like BLIP, LLaVA, and InstructBLIP \citep{li2022blip,dai2023instructblip,ye2023mplug,liu2024visual}. These models leverage image-text pretraining to achieve sophisticated multimodal understanding, with recent advances incorporating efficient pseudo-labels for complex visual reasoning tasks, such as Kosmos-2 and PaLI-X \citep{peng2023kosmos,chen2023pali}.


Despite these advances, LVLMs still face significant challenges in real-world applications, particularly in domain-specific tasks and cross-scenario generalization. Researchers have explored various enhancement strategies, from incorporating expert knowledge to developing targeted data augmentation techniques \cite{xiong2024llava,zhang2024critic}. These ongoing challenges underscore the critical importance of improving LVLMs' multimodal comprehension and domain robustness \cite{zhang2024internlm,zhang2024spa}.

\subsection{Data Augmentation in LVLMs}
While data augmentation has proven successful in computer vision tasks, its application to LVLMs reveals unexpected challenges and insights \citep{chen2020simple,grill2020bootstrap,he2020momentum,zhu2023multi}. Recent studies show that LVLMs exhibit remarkable sensitivity to minor image perturbations—even simple transformations like random flipping can induce significant semantic shifts in model outputs. This behavior, distinctly different from traditional computer vision models' robustness to augmentation, suggests fundamental differences in how LVLMs process and interpret visual information. Standard augmentation strategies from contrastive learning often lead to inconsistent model behaviors rather than improved robustness.

Intriguingly, this sensitivity to visual perturbations has emerged as a valuable property for model improvement \citep{awais2025foundation,yuan2024self,chen2024self,yu2023prevent}. The semantic divergence between outputs from original and augmented inputs creates informative preference pairs, offering a novel approach to model alignment without extensive manual annotation.


\begin{figure*}[t]
    \centering
    \includegraphics[width=\linewidth]{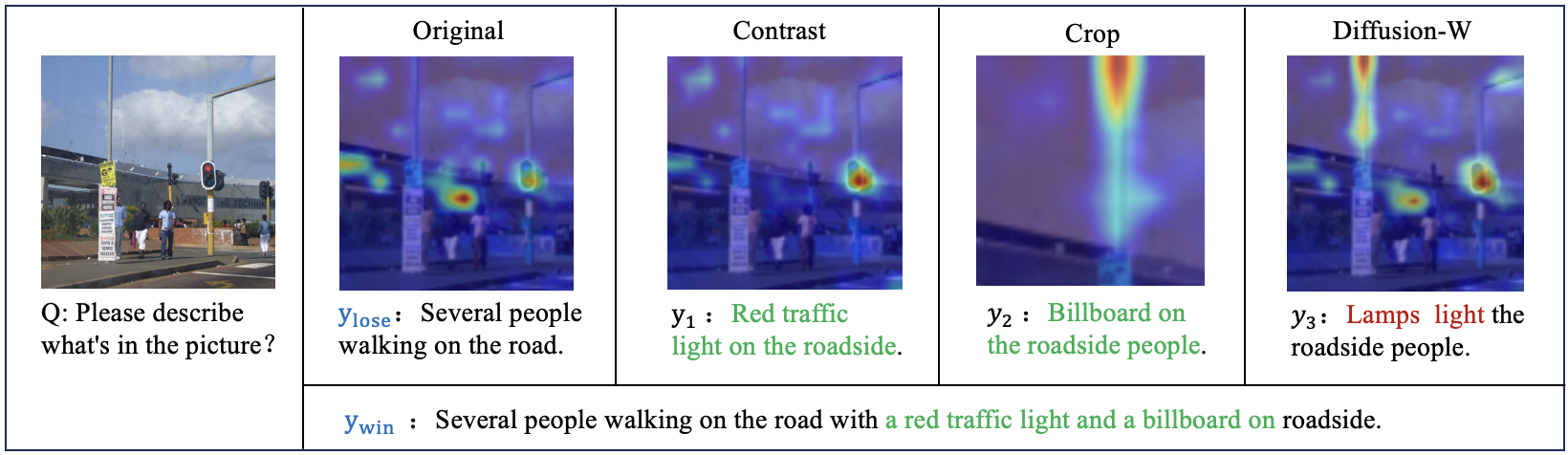}
    \caption{The winner answer $y_{win}$ through attention visualization indicates how SHAPE enables holistic caption generation. And $y_{lose}$ is the original generation. The \textcolor{darkgreen}{green} text represents correctly recognized content, while the \textcolor{darkred}{red} text represents incorrect recognition.}
    \label{fig:att}
\end{figure*}

\subsection{LVLMs Preference Alignment}
Aligning LVLM outputs with human preferences remains a critical challenge, primarily due to the independent pre-training of visual and language modalities. While conventional methods like PPO and DPO have demonstrated effectiveness in language models \citep{schulman2017proximal, rafailov2024direct,li2025naturalbench}, their application to multimodal scenarios faces key limitations: they require costly human preference data, depend heavily on external AI feedback, and often overlook models' inherent capabilities.

These challenges are amplified in multimodal contexts, where precise visual-linguistic alignment is crucial. Current approaches largely overlook intrinsic LVLMs characteristics, particularly their sensitivity to visual perturbations \cite{tong2025cambrian,diao2025unveiling,yang2025opadpo}, which could serve as valuable alignment signals. Although recent work has begun exploring model-inherent properties, the potential of leveraging visual sensitivity for preference pair generation remains untapped. This gap motivates the development of efficient, model-specific alignment strategies that can fully utilize these unique characteristics.


\section{Method}
\label{sec:method}


\noindent \partitle{Motivation} Inspired by these observations detailed in Figure \hyperref[fig:att]{2} and Table \hyperref[tbl:aug_dpo]{3}, our motivation lies in transforming what was once considered a limitation—the model's susceptibility to input variations—into an opportunity. We leverage these variations as an automated supervision mechanism, thus optimizing alignment processes, promoting semantic richness, and maintaining low annotation costs while enhancing the comprehensive understanding capabilities of LVLMs.

\begin{figure*}[!t]
    \centering
    \includegraphics[width=0.92\linewidth]{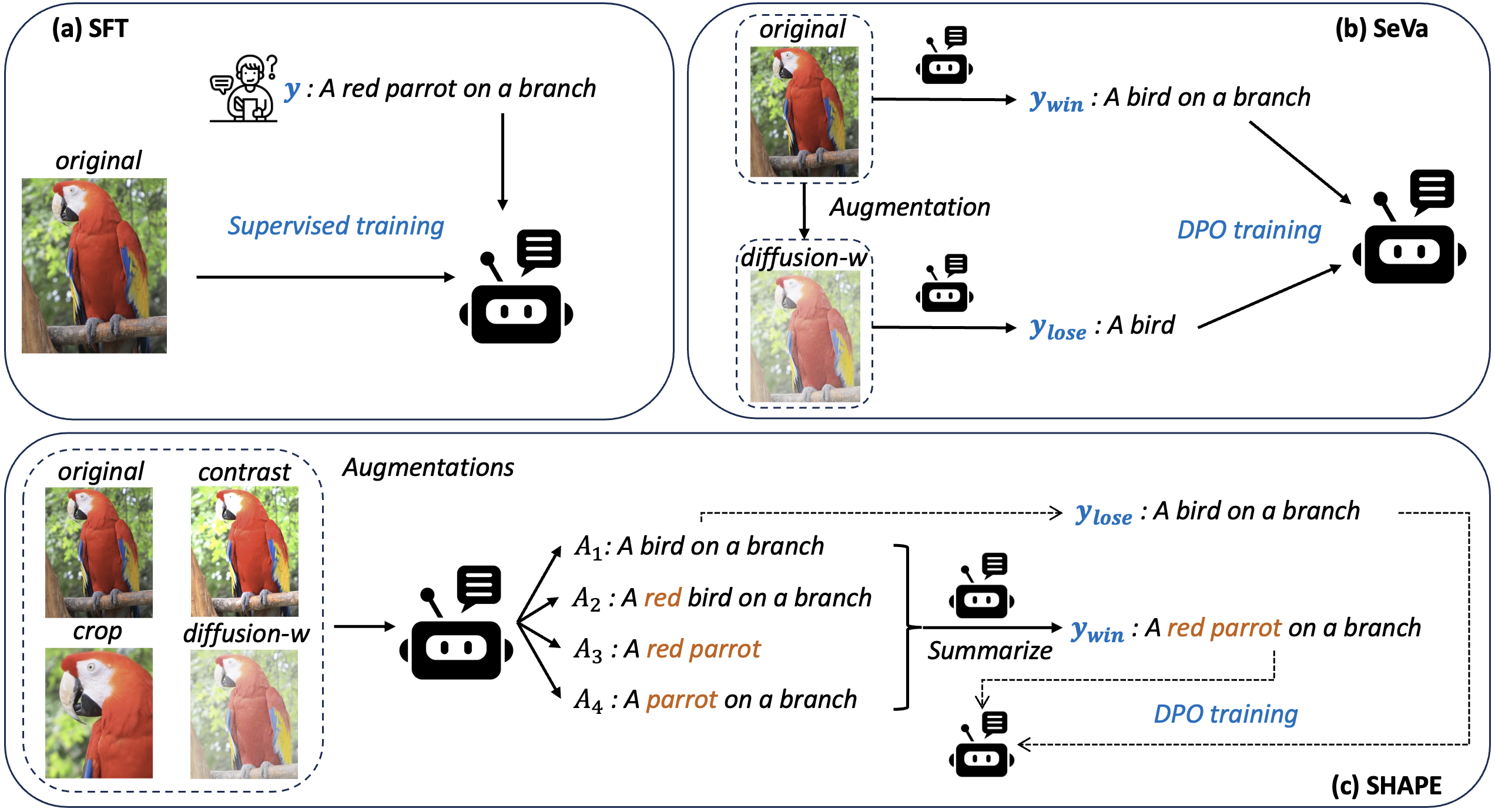}
    \caption{Overview of \projectname: Unlike traditional SFT, which relies on single-path supervised learning with human preference intervention (e.g., (a) SFT), or simple augmentation methods (e.g., (b) SeVa), \projectname~ fully leverages the potential of LVLMs by extending the self-supervised optimization paradigm to Visual Question Answering (VQA). It enriches image-side understanding to generate holistic training signals, enabling more reliable and detailed visual comprehension without requiring additional annotations.}
    \label{fig:framework}
\end{figure*}

We present \projectname~, a framework that transforms standard supervised text-image pairs into more holistic preference triplets for LVLMs alignment. Instead of relying on costly preference annotations, our approach leverages existing supervised data to construct competitive preference pairs for efficient self-supervised preference learning. As shown in Figure \hyperref[fig:framework]{3}, \projectname~ extends beyond traditional SFT (Figure \hyperref[fig:framework]{3(a)}) and simple augmentation methods like SeVa (Figure \hyperref[fig:framework]{3(b)}), by implementing multiple image augmentations to generate diverse perspectives that are synthesized into more comprehensive outputs (Figure \hyperref[fig:framework]{3(c)}).

\partitle{Preference Optimization}
Preference optimization has shown promise in fine-tuning language models and aligning their behavior with desired outcomes. Given an input prompt $x$, a language model with policy $\pi_\theta$ can produce a conditional distribution: 
\begin{equation*}
    \pi_\theta(y|x) \gets \Pi_{k=1}^L \pi_\theta(y_k | y_{<k},x),
\end{equation*}
where $y_{<k}$ represents the generated token before the current prediction
$y_k$, and $L$ is the length of the token sequence.

The preference data is defined as $\mathcal{D}=\{(x^i, y^i_w, y^i_l)\}_{i=1}^N$, where $y^i_w$ denotes the preferred response for the input prompt $x^i$, and $y^i_l$ denotes the dispreferred one. Taking DPO \cite{rafailov2024direct} as a representative example, it formulates the
probability of obtaining each preference pair as a Bradley-Terry (BT) preference model:
\begin{equation*}
    p(y_w \succ y_l|x) = \frac{\exp(r(x, y_w))}{\exp(r(x, y_w)) + \exp(r(x, y_l))},
\end{equation*}
where $r(x, y)$ represents the reward or quality score for answer $y$ given input $x$. This formulation naturally captures our intuition that the winning answer should have a higher probability of being preferred, while maintaining a meaningful comparison with the competitive loser. DPO optimizes the language models with the following loss: 
{\small
\begin{equation*}
    \mathcal{L}(\pi_\theta; \pi_\text{ref}) = -\mathbb{E}_{\mathcal{D}}\log \sigma \left( \beta \log \frac{\pi_{\theta}(y_w|x)}{\pi_{\text{ref}}(y_w|x)} - \beta \log \frac{\pi_{\theta}(y_l|x)}{\pi_{\text{ref}}(y_l|x)} \right),
\end{equation*}
}

\partitle{Self-supervised Preference Optimization}
In SeVa \cite{zhu2024self}, they construct the preference dataset with the original image's output as the winner choice and the augmented image's output as the loser choice:
\begin{align*}
    y_w &\gets \pi_\theta(\cdot|x), \\
    y_l &\gets \pi_\theta(\cdot|f(x)),
\end{align*}
where $x$ is the input of LVLMs, i.e. the question concatenated with image. $f(\cdot)$ is the data augmentation method, i.e. diffusion noise augmentation.

We note that for the same image, the output of model reasoning is different when different image-augmented methods are used. Although they may focus on fewer details, the diversity is improved. Figure \ref{fig:att} shows how the attention patterns of LVLMs change with different data augmentation methods, resulting in a broader range of candidate outputs. Based on this observation, We can summarize the outputs of different image-augmented methods as the winner choice and set the origin output as loser choice:
\begin{align*}
    y_l &\gets \pi_\theta(\cdot|x), \\
    y^i &\gets \pi_\theta(\cdot|f_i(x)), \quad i\in[1,...,M] \\
    y_w & \gets \pi_\theta(\cdot|y^1|| y^2||...||y^M|| p),
\end{align*}
where $p$ is the summarization prompt, such as ``Please provide a comprehensive summary based on the following candidate answers." 

After obtaining the preference data, we fine-tune the aligned visual model using DPO. For each iteration $t$, we use the last iteration fine-tuned model $\pi_{\theta_{t-1}}$ as the reference model. Then the loss at iteration $t$ is:
{\small
\begin{equation*}
    \mathcal{L}_t = -\mathbb{E}_{\mathcal{D}}\log \sigma \left( \beta \log \frac{\pi_{\theta_t}(y_w|x)}{\pi_{\theta_{t-1}}(y_w|x)} - \beta \log \frac{\pi_{\theta_t}(y_l|x)}{\pi_{\theta_{t-1}}(y_l|x)} \right).
\end{equation*}
}

 We provide the workflow of \projectname~ in Algorithm \ref{alg:shape}. The procedure iteratively refines the visual language model through self-supervised preference optimization.

\begin{algorithm}[H]
\caption{Self-supervised Visual Preference Alignment}
\begin{algorithmic}[1]
\label{alg:shape}
\REQUIRE Datasets: $\mathcal{D}=\{x^i\}_{i=1}^N$; Initial Model: $\pi_{\theta_0}$; Image augmentation methods $\{f_j(\cdot)\}_{j=1}^M$; Iterations: $T$.
\STATE $\pi_{\text{ref}} \gets \pi_{\theta_0}$ 
\hspace{0.8cm}// Initialize the reference model.
\STATE $\mathcal{D}_p = \{\}$
\hspace{1cm} // Initialize the preference dataset.
\FOR{$t = 1,...,T$}
    \FOR{each $x^i \in \mathcal{D}$}
        \FOR{each augmentation methods $f_j(\cdot)$}
            \STATE $x^i_j \gets f_j(x^i)$
            \hspace{0.5cm} // Augment images.
            \STATE $y^i_j \sim \pi_{\text{ref}}(\cdot | x^i_j)$
            \hspace{0.2cm} // Get candidate answers.
        \ENDFOR
        \STATE $y^i_l \gets \pi_{\theta}(\cdot | x^i)$
        //Take origin output as loser choice. \\
        \STATE //Summarize candidate answers as winner choice.
        \STATE $y^i_w \gets \textrm{Summarize}(y^i_1, y^i_2, ..., y^i_M)$
        \STATE $\mathcal{D}_p \gets \mathcal{D}_p \cup \{x^i, y^i_w, y^i_l\}$ //Update dataset.
    \ENDFOR
    \STATE $\pi_{\theta_t} \gets \text{argmin}_\theta \mathcal{L}_t(\pi_{\theta_t}, \pi_{\text{ref}})$ //Update target model.
    \STATE $\pi_{\text{ref}} \gets \pi_{\theta_t}$
    \hspace{0.5cm} //Update reference model.
\ENDFOR
\ENSURE Aligned visual model $\pi_{\theta_T}$.
\end{algorithmic}
\end{algorithm}


\begin{table*}[!t]
        \renewcommand\arraystretch{1.1}
	\centering
	\begin{tabular}{l|ccccccc|cc|ccccc}
		\toprule[1pt]
		\multirow{2}{*}{\textbf{Model}}  & \multicolumn{7}{c|}{\textbf{MMVet}}& \multicolumn{2}{c|}{\textbf{MMBench}} & \multicolumn{4}{c}{\textbf{POPE}} \\

      & \textbf{All} & rec  & ocr &know & gen & spat &  math & en & cn & \textbf{All} & rand & pop & adv\\
		\midrule[1pt]
        LLaVA-1.5-7B &  30.5 &  35.7 &  21.9 &  17.7 &  19.7 &  24.7 &  \textbf{7.7} &  64.3 &  58.3 &  85.9 &  89.5 &  86.7 &  81.7\\
        \quad + CSR & 33.9 & 39.2 & 23.3 & 21.9 & 24.5 & 27.7 & \textbf{7.7} & 65.5 & \textbf{59.4} & 86.8 & 89.4 & 87.4 & 83.6 \\
        \quad + SeVa & 37.2 & 40.2 & 29.9 & 21.8 & 23.9 & 34.3 & \textbf{7.7} & 65.6 & 59.2 & 86.7 & 89.4 & 87.1 & 83.6 \\
        \quad + \projectname & \textbf{\textcolor{orange}{41.8}} & \textbf{46.1} & \textbf{33.8} & \textbf{32.5} & \textbf{37.0} & \textbf{39.2} & \textbf{7.7} & \textbf{\textcolor{orange}{65.7}} & \textbf{\textcolor{orange}{59.4}} & \textbf{\textcolor{orange}{93.9}} & \textbf{95.5} & \textbf{95.2} & \textbf{91.0}\\
        \midrule[1pt]
         LLaVA-1.5-13B &  35.4 & 38.9 & 32.2 & 23.3 & 24.8 & 29.7 & 24.8 & 67.7 & 63.6 & 85.9 & 89.6 & 86.5 & 82.0\\
        \quad + CSR & 37.8 & 41.0 & 32.5 & 24.6 & 30.1 & 32.8 & 24.8 & 68.8 & 64.5 & 87.3 & 90.4 & 89.1 & 82.2 \\
        \quad + SeVa &  41.0 &  45.4 &  32.8 &  32.4 &  36.7 &  37.0 &  \textbf{25.4} &  68.7 &  64.8 &  87.4 &  90.5 &  89.0 &  82.7 \\
        \quad + \projectname  & \textbf{\textcolor{orange}{42.8}} & \textbf{46.7} & \textbf{37.7} & \textbf{34.6} & \textbf{39.0} & \textbf{39.7} & 25.0 & \textbf{\textcolor{orange}{69.0}} & \textbf{\textcolor{orange}{65.1}} & \textbf{\textcolor{orange}{94.5}} & \textbf{96.0} & \textbf{95.9} & \textbf{93.3}\\
		\bottomrule[1pt]
	\end{tabular}
\caption{Comparison of \projectname~ with state-of-the-art enhancement strategies applied to the LLaVA-1.5-7B (in both 7B and 13B model sizes) across the MMVet, MMBench, and POPE benchmarks.}
\label{tab:vlm_compare1}
\end{table*}


\section{Experiment}
\label{sec:experi}

\subsection{Settings}
\noindent \partitle{Data construction} 
The source data are obtained from the LLaVA665k SFT dataset \cite{liu2024improved}. Specifically, image-question pairs from TextVQA \cite{singh2019towards} and OCRVQA \cite{mishra2019ocr} (collectively referred to as ``text+ocr") within LLaVA665k are used to generate the DPO preference data.

\partitle{Training Strategies} 
Following the settings of prior work \cite{zhao2023beyond,liu2024improved}, we take CLIP-VIT-L-336px as vision encoder, the batch size is 128, and the learning rate is $2e^{-6}$. The default LoRA rank $r$ is set to 1024 and the scale parameter $\beta$ in DPO is fixed at 0.1.

\partitle{Evaluation Benchmarks} 
We evaluate \projectname~ across three categories of benchmarks: comprehensive benchmarks, general VQA and hallucination benchmarks. There includes: (1) Comprehensive benchmarks(MM-Vet \cite{yu2023mm}, MMbench \cite{liu2025mmbench}, MME \cite{yin2023survey}, SEED \cite{li2023seed}, LLaVA-Bench \cite{liu2024visual}); (2)General VQA (ScienceQA \cite{lu2022learn}, GQA \cite{hudson2019gqa}, VisWiz \cite{gurari2018vizwiz}); (3) Hallucination benchmark(POPE \cite{li2023evaluating}, CHAIR \cite{rohrbach2018object})
. More detailed descriptions are provided in Appendix \ref{benchmark_detail}.

\partitle{Data Augment Impact} \label{part:aug_impact}
After an extensive exploration of data augmentation strategies and careful consideration of summary elements, we chose a combination of three augmentation strategies along with the original image. We evaluated various data augmentation combinations to optimize self-supervised training process of \projectname~ for LVLMs. Based on comprehensive comparative experiments in Table \ref{tbl:aug_dpo}, we identified an optimal set of augmentation strategies, focusing on transformations in three key dimensions of the image: contrast, noise, and spatial region. These selected strategies were integrated into the training pipeline of \projectname~ to maximize data diversity.

A critical enhancement to our method was the addition of a response summarization step. By consolidating multiple augmentation-generated perspectives into a cohesive response, this step bridges the gap between diverse individual outputs and comprehensive understanding. This approach forms the foundation for our experimental design in the subsequent ablation studies.

\begin{table*}
        \renewcommand\arraystretch{1.1}
	\centering
	\begin{tabular}{lccccccccc}
		\toprule[1pt]
	\textbf{Model} & $\textbf{MME}^\text{P}$ & $\textbf{MME}^\text{C}$ & \textbf{SEED} & $\textbf{LLaVA}^\text{W}$ & \textbf{SQA} & \textbf{GQA} & \textbf{VisWiz} & $\textbf{CHAIR}_\text{s}\downarrow$ & $\textbf{CHAIR}_\text{i}\downarrow$\\
    \midrule[1pt]
     LLaVA-1.5-7B & 1510.7 & 348.2 & 58.6 & 63.4  & 66.8 & 62.0 & 50.0 & 48.8 & 14.9 \\
     \quad + CSR & 1524.2 & 367.9 & 60.3 & 71.1 & 70.7 & 62.3 & \textbf{54.1} & 21.0 & 6.0 \\
     \quad + SeVa & 1531.0 & 369.2 & 65.8 & 72.2 & 67.5 & 60.7 & 51.5 & 20.5 & 5.8 
     \\
     \quad + \projectname & \textbf{1539.2} & \textbf{370.0} & \textbf{67.4} & \textbf{73.1} & \textbf{71.5} & \textbf{62.8} & 55.3 & \textbf{19.5} & \textbf{5.4} \\
		\midrule[1pt]
  LLaVA-1.5-13B & 1531.3 & 295.4 & 61.6 & 70.7 & 71.6 & 63.3 & 53.6 & 48.3 & 14.1 \\
     \quad + CSR & 1530.6 & 303.9 & 62.9 & 74.7 & 75.1 & 63.7 & \textbf{56.8} & 28.0 & 7.3 \\
     \quad + SeVa & 1533.9 & 305.1 & 68.6 & 80.1 & 71.2 & 63.4 & 54.7 & 23.6 & 6.5 
     \\
     \quad + \projectname & \textbf{1544.0} & \textbf{306.9} & \textbf{69.3} & \textbf{82.5} & \textbf{72.5} & \textbf{64.4} & 55.8 & \textbf{23.1} & \textbf{6.6} \\
     \bottomrule[1pt]
	\end{tabular}
\caption{Comparison of \projectname~ with state-of-the-art enhancement strategies applied to the LLaVA-1.5-7B across various evaluation benchmarks. \projectname~ achieved the best scores on most tasks. In the CHAIR tasks, where lower scores are better, \projectname~ also showed better performance, indicating that it effectively improved the overall performance of the model.}
\label{tab:vlm_compare2}
\vspace{-1.5em}
\end{table*}

\subsection{Comparison with State of the Art}
To comprehensively evaluate the effectiveness of our proposed alignment strategy, we compare \projectname~ against several SOTA models, including LLaVA-1.5, CSR, and SeVa. The results indicate that \projectname~ consistently outperforms these baselines across a range of evaluation benchmarks, demonstrating its superior ability in preference alignment, robustness, and task-specific capabilities.

We selected these baselines for their distinct approaches to preference alignment in LVLMs, which have been pivotal in advancing large vision-language models without relying heavily on manually labeled data. Specifically, LLaVA-1.5 serves as a foundational baseline that lacks advanced alignment techniques, providing a point of reference for evaluating the value added by different alignment strategies. CSR utilizes a reward-based calibration mechanism, attempting to form preference pairs through a self-reinforcing feedback loop that effectively reduces the reliance on annotated datasets. SeVa implements DPO but primarily compares raw and enhanced outputs, limiting its capacity for capturing semantic nuances. These diverse strategies provide a comprehensive backdrop for analyzing the advantages brought by \projectname~, particularly in leveraging unsupervised data to align model outputs effectively and overcome the existing limitations in preference alignment.

\partitle{\projectname~ Bootstraps LVLM's Performance} 
The results in Table \ref{tab:vlm_compare1} and Table \ref{tab:vlm_compare2} illustrate that \projectname~ consistently outperforms baseline models across multiple benchmarks, highlighting its strengths in various vision-language tasks.

Across MMVet and MMBench (shown in Table \ref{tab:vlm_compare1}), \projectname~ achieved superior overall scores regardless of model size. In the 7B setting, it attains an ``All" score of 41.8 on MMVet, surpassing SeVa (37.2) and CSR (33.9). This trend continues in the 13B setting, where \projectname~ maintains its lead with an ``All" score of 42.8. The consistently better performance across different scales suggests that the proposed alignment strategy is not only effective but also scalable, offering robust enhancements to the model’s comprehension abilities as capacity increases.

Task-specific analysis reveals strengths in knowledge-based and generative tasks, with our method achieving scores of 34.6 and 39.0 on MMVet's respective subtasks in the 13B setting, surpassing all baselines including in general VQA evaluations. The approach demonstrates improved accuracy and reduced hallucinations as evidenced by POPE and CHAIR evaluations, suggesting it effectively addresses current alignment limitations across diverse vision-language tasks while generating more informative and contextually appropriate responses.

\begin{figure*}[!t]
    \centering
    \includegraphics[width=\linewidth]{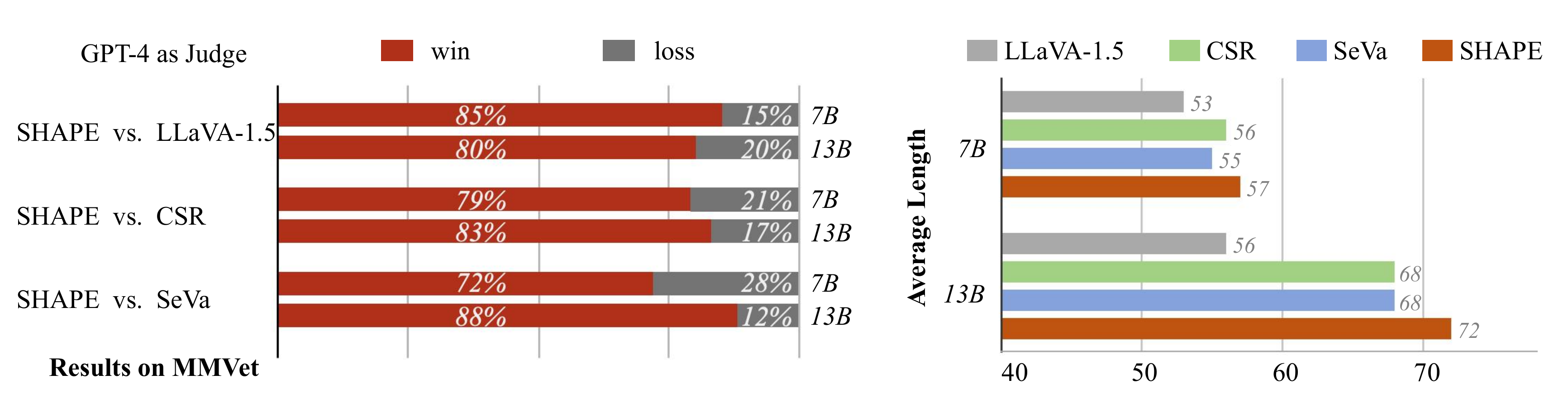}
    \caption{Evaluation of \projectname~, SeVa, CSR, and LLaVA-1.5 (7B and 13B) on MMVet, comparing win rates and average output lengths, with GPT-4 as the judge for visual-language task performance.}
    \label{fig:winrate}
\end{figure*}

\partitle{Why \projectname~ Better} 
Figure \ref{fig:winrate} presents a comparative evaluation of SHAPE, SeVa, CSR, and LLaVA-1.5-7B/13B models on the MMVet benchmark, with GPT-4V serving as the judge. This analysis examines two key metrics: pairwise win rates and average response length. The GPT-4V evaluation offers insights into each model’s effectiveness in visual-language tasks. For details on GPT-4V-assisted preference annotation, refer to Appendix \ref{sec:gptjudge}.

Win rate analysis suggests that \projectname~ generates more informative responses than LLaVA-1.5, CSR, and SeVa across both 7B and 13B variants. On average, it produces 57-token responses in the 7B configuration and 72 tokens in the 13B configuration, whereas other models generate shorter outputs. These findings align with prior research \cite{zhu2023minigpt,sun2024sq,shin2024x}, which associates longer responses with improved multimodal comprehension. This trend is more pronounced in larger models, potentially explaining \projectname~’s increased win rate against SeVa from 72\% to 88\% when scaling from 7B to 13B. Qualitative comparisons are available in Appendix \ref{shape_case}.

\subsection{Ablation Study}
To gain deeper insights into the effectiveness of \projectname~ in aligning LVLMs, we conduct ablation studies to isolate the impact of key components, including data augmentation, summarization, and architectural parameters, identifying the optimal configuration.



\partitle{Data Augmentation Strategies}
Building on our framework from paragraph \ref{part:aug_impact}, we conducted detailed experiments to quantify the impact of specific augmentation combinations. We evaluated both individual augmentation strategies and their combinations to identify the most effective configuration, assessing performance on MMVet, MMBench, and POPE benchmarks.

Experimental results indicate that moderate augmentation strength yields the best performance, as both overly strong and excessively weak augmentations negatively affect DPO training. Based on the experimental results and the suitability of the augmentation strength as detailed in Appendix \ref{sec:strategies_impact}, we finally selected candidate-3 (Contrast + Diffusion-W + Gamma) as our augmentation scheme, wherein each augmentation was applied to distort images (along with the questions) to produce ``rejected" responses, paired with ``chosen" responses from the original image input. This approach enabled the creation of high-quality preference pairs, subsequently used to guide the DPO training process.

Evaluation results further confirmed the benefits of these augmentation strategies for multimodal comprehension, helping the model more clearly distinguish between ``chosen" and ``rejected" responses.  The progressive strategy combination analysis reveals the importance of strategic integration, as shown in Table \ref{tbl:aug_dpo}. While individual augmentations provide modest gains, their combination yields substantially stronger results. Specifically, the candidate-3 configuration provided the optimal balance between performance gains and complexity. Integrating multiple strategy combinations significantly enhanced the extraction of local detail features, resulting in superior summarization capabilities.

\todo{}
\begin{table}[!t]
\small
\renewcommand\arraystretch{1.0}
\centering
\setlength{\tabcolsep}{8pt}
\begin{tabular}{lccc}
\hline
\textbf{Method} & \textbf{MMVet} & \textbf{MMB} & \textbf{POPE} \\
\hline
LLaVA-1.5-7B  & 30.5 & 64.3 & 85.9 \\
\quad+ RandFlip & 33.7 & 64.4 & 86.7\\
\quad+ Contrast & 35.6 & 65.1 & 86.7\\
\quad+ Crop(0-20\%) & 35.0 & 64.5 & 86.1\\
\quad+ Diffusion-S  & 34.6 & 65.2 & 86.6\\
\quad+ Diffusion-W  & \textbf{37.2} & \textbf{65.6} & \textbf{89.4}\\
\quad+ Gamma & 37.0 & 65.2 & 89.2\\
\hline
\projectname~ \\
\quad+ Candidate-1 & 38.7 & 65.2 & 87.4\\
\quad+ Candidate-2 & 41.4 & 65.4 & 89.4\\
\quad+ Candidate-3 & 41.7 & 65.7 & 92.7\\
\quad+ Candidate-4 & 41.8 & 65.9 & 93.6\\
\hline
\end{tabular}
\caption{Comparison of augmentation strategies to identify the best-performing schemes on MMVet, MMB and POPE. The top section shows individual augmentation results, the middle section shows performance with varying numbers of combined strategies, and the bottom section shows results with different summary combinations. And SeVa strategies use Diffusion-W on the LLaVA-1.5-7B.}
\label{tbl:aug_dpo}
\end{table}

\partitle{Impact with Summary Module} 
\projectname~ enhances preference alignment through its advanced summarization capability, generating richer and more context-aware summaries. To assess the impact of this summarization functionality, we compare \projectname~ with SeVa, a model with a similar architecture but without the summarization feature. As shown in Table \ref{tbl:sft_dpo}, while SeVa achieves competitive results, with a score of 37.2 on MMVet and 65.6 on MMB (using 16k dataset scale), \projectname~ consistently outperforms it, achieving 41.8 on MMVet and 65.7 on MMB with the same data scale. This clearly demonstrates that the summarization feature in \projectname~ significantly improves alignment quality and overall model comprehension. 

The analysis further examined three specific summary configurations: summary-1 (Contrast + Diffusion-W + Gamma), summary-2 (Crop + Diffusion-S + Gamma), and summary-3 (RandFlip + Diffusion-S + Diffusion-W). Table \ref{tbl:aug_dpo} shows summary-1 performed best on MMVet and POPE, while summary-2 slightly outperformed on MMB. These results indicate that different augmentation combinations offer varying benefits across benchmarks, with the optimal configuration depending on the specific characteristics of the target task. The summarization mechanism provides more nuanced guidance during preference optimization, enabling \projectname~ to excel in tasks requiring deep contextual understanding.

\partitle{Impact with Target Answers} 
To assess the advantage of \projectname~’s unsupervised alignment strategy, we compare it with SFT using subsets of LLaVA665k—specifically 15k (2\%), 66k (10\%), and 132k (20\%) instances. Similar to SeVa, \projectname~ significantly reduces dependency on labeled data while still achieving superior performance. \projectname~ reaches a score of 41.8 on MMVet using only 16k unsupervised samples, surpassing SFT performance even when using larger labeled datasets of up to 132k instances, as detailed in Table \ref{tbl:sft_dpo}. This highlights the effectiveness of \projectname~'s unsupervised preference optimization, which not only minimizes reliance on costly annotations but also delivers higher performance with significantly less data and training effort.

\begin{table}[!t]
\small
\renewcommand\arraystretch{1.2}
\centering
\setlength{\tabcolsep}{8pt}
\begin{tabular}{lcccc}
\hline
\textbf{Method} & \textbf{Scale} & \textbf{MMVet} & \textbf{MMB} & \textbf{POPE} \\
\hline
LLaVA-1.5 & - & 30.5 & 64.3 & 85.9 \\
\quad- SFT(10\%) & 66k & 32.8 & 64.9 & 86.0\\
\quad- SFT(20\%) & 66k & 33.9 & 64.2 & 86.1\\
\quad- SFT & 102k & 32.5 & 65.2 & 86.7\\
\hline
SeVa & 8k & 34.8 & 65.3 & 86.2\\
SeVa & 16k & 37.2 & 65.6 & 86.7\\
\hline
\projectname & 8k & \textbf{36.5} & \textbf{65.5} & \textbf{90.3}\\
\projectname & 16k & \textbf{41.8} & \textbf{65.7} & \textbf{93.6}\\
\hline
\end{tabular}
\caption{Comparison of SFT, SeVa, and \projectname~ on LLaVA-1.5-7B across MMVet, MMB, and POPE benchmarks. SFT uses subsets of LLaVA665k (15k, 66k, and 102k samples), while SeVa and \projectname~ rely on unsupervised data without ground truth annotations. \projectname~ achieves superior performance with smaller data subsets and reduced training costs, highlighting its efficiency.}
\label{tbl:sft_dpo}         
\end{table}

the performance of SFT even with larger labeled datasets, outperforming SFT even when larger labeled datasets are used, as detailed in Table \ref{tbl:sft_dpo}. 

\begin{table}[!t]
\small
\centering
\setlength{\tabcolsep}{8pt}
\renewcommand\arraystretch{1.2}
\begin{tabular}{ccccc}
\hline

\textbf{LoRA $r$} & \textbf{Method} & \textbf{MMVet} & \textbf{MMB} & \textbf{POPE} \\
\hline
- & LLaVA-1.5 & 30.7 & 64.3 & 85.9 \\
\hline
\multirow{2}{*}{256} & SeVa & 34.0 & 65.1 & 86.6 \\
& \projectname & 38.7 & 64.9 & 91.0\\
\hline
\multirow{2}{*}{512} & SeVa & 35.5 & 65.5 & 86.8 \\
& \projectname & 41.2 & 65.6 & 93.2 \\
\hline
\multirow{2}{*}{1024} & SeVa & 37.2 & 65.6 & 86.7 \\
& \projectname & 41.8 & 65.7 & 93.6 \\
\hline
\multirow{2}{*}{2048} & SeVa & 33.5 & 65.0 & 84.5 \\
& \projectname & 40.5 & 65.2 & 93.0 \\
\hline
\end{tabular}
\caption{Performance comparison across LoRA \cite{hu2021lora} rank $r$ from 256 to 2048. \projectname~ achieves consistently strong results, with optimal performance at $r$=1024.}
\label{tbl:lora_dpo}
\end{table}

\partitle{Impact of LoRA Module} 
To investigate the influence of the LoRA adaptation module, we conducted an ablation study with varying low-rank hyper-parameters across three multimodal comprehension benchmarks, as shown in Table \ref{tbl:lora_dpo}. The parameter choice is crucial because it determines how much new knowledge the model can effectively absorb during fine-tuning. We found that increasing the rank $r$ generally boosts performance on downstream tasks, suggesting that a higher rank allows the model to capture more knowledge. However, further increasing $r$ to 2048 triggered performance degradation(40.5 on MMVet, 65.2 on MMB), consistent with catastrophic forgetting during the training of the LLM \cite{fu2024dtl}. Consequently, \projectname~ adopted $r$=1024 to achieve the best trade-off, with the delta importance factor $\alpha$ consistently set at twice the value of $r$, ensuring a stable and effective fine-tuning process.

\begin{table}[!h]
\centering
    \resizebox{0.48\textwidth}{!}{
    \large
    \begin{tabular}{lcccc}
        \toprule
       \textbf{Method} & \textbf{Offline (h)} & \textbf{Training (h)} & \textbf{Total (h)} &  \textbf{MMVet} \\
        \midrule
        SeVa & 0.05 & 3.50 & 3.55 & 37.2   \\
        SHAPE & 0.21 & 3.50 & 3.71 & 42.8  \\
        \bottomrule
    \end{tabular}
    }
\caption{Training cost/result of LLaVA-1.5-7B on 16k samples.} 
\label{tbl:training}
\end{table}

\partitle{Trade-off between training cost and efficiency} 
While \projectname~ necessitates additional computational resources for offline data generation, although these constitute a modest proportion of the total duration of training. Table \ref{tbl:training} shows the self-augmented training results of LLaVA-1.5-7B on 4 A100 GPUs with 16k training data (8k TextVQA + 8k OCRVQA). SHAPE only has 9 min more offline time than SeVa, but it brings a 5.6\% improvement to MMVet. Overall, the trade-off in time for offline data augmentation is worthwhile.

\section{Conclusion}
\label{sec:conclusion}

In this paper, we have proposed \projectname, a self-supervised framework tailored for Large Visual Language Models to achieve more effective preference alignment without the need for human preference annotations. \projectname~ has achieved this by transforming off-the-shelf text-image pairs datasets into preference data, where each preference triplet consists of a winner text that consistently improves in holisticness and outperforms the loser text in quality, thereby pushing the model to progressively enhance the alignment performance throughout preference finetuning. The extensive experiments have demonstrated that \projectname~ significantly enhances performance across multiple benchmarks, showing consistent gains for both 7B and 13B model sizes. Our qualitative analysis has further highlighted that LVLMs yielded from \projectname~ exhibit enhanced attention to visual details and provide richer, contextually aware responses aligned with human preferences.

\paragraph{Limitations and Future Works.} 
\projectname~ still relies on a predefined strategy for visual augmentations, which may limit the diversity of transformations. Additionally, we plan to investigate automated prompt generation and refinement to better adapt to different data distributions, thereby improving the quality of the text summarization step, which is currently designed manually. 
Further validation across diverse real-world tasks is also required to assess generalizability more comprehensively.

{
    \small
    \bibliographystyle{ieeenat_fullname}
    \bibliography{main}
}

\clearpage
\newpage
\maketitlesupplementary

\section{Appendix}
\label{sec:appendix}
This appendix provides supplementary details on methodology and experimental procedures. We begin with a description of the evaluation benchmarks \ref{benchmark_detail} used to assess model performance. Next, we elaborate on the experimental setup \ref{sec:strategies_impact}, including the choice of image augmentation strategies and their impact on model performance. Finally, we present qualitative results \ref{shape_case}, including visual comparisons and GPT-4V annotations \ref{sec:gptjudge}, which demonstrate the model’s effectiveness in aligning visual and textual information. Source code is publicly available at \url{https://anonymous.4open.science/r/shape-002C}.

\subsection{Evaluation Benchmarks}
\label{benchmark_detail}

\underline{LLaVA-Bench (In the wild)} \cite{liu2024visual}: A challenging benchmark of 60 diverse tasks designed to evaluate models in naturalistic settings. It specifically tests visual instruction-following and question-answering capabilities in real-world scenarios, offering insights into practical applicability.

\underline{MM-Vet} \cite{yu2023mm}: A comprehensive evaluation suite comprising 128 diverse tasks that assess six core visual-language capabilities. This benchmark uniquely combines mathematical reasoning, logical inference, and visual knowledge understanding, providing a rigorous test of multi-modal comprehension.

\underline{MM-Bench} \cite{liu2025mmbench}: A large-scale multi-modal benchmark with 4.7K samples, focusing on visual knowledge and reasoning capabilities. This dataset provides a balanced assessment of both factual knowledge and analytical reasoning in multi-modal contexts.

\underline{POPE} \cite{li2023evaluating}: A specialized benchmark containing 8,440 samples designed to evaluate model hallucination. It specifically tests models' ability to provide accurate Yes/No responses about object presence in images, serving as a critical measure of visual grounding reliability.

\underline{MME} \cite{yin2023survey}: A benchmark with 14 tasks assessing perception and cognition in LVLMs, challenging interpretative and analytical skills.

\underline{SEED} \cite{li2023seed}: A benchmark designed to evaluate the generative comprehension capabilities of large vision-language models (LVLMs). It includes an extensive dataset of 19K multiple-choice questions with precise human annotations, spanning 12 distinct evaluation dimensions that cover both spatial and temporal understanding across image and video modalities.

\underline{ScienceQA} \cite{lu2022learn}: A multimodal benchmark crafted to evaluate and diagnose the multi-hop reasoning abilities and interpretability of AI systems within the science domain. It features an extensive dataset of approximately 21k multiple-choice questions, spanning a broad spectrum of scientific topics and supplemented with detailed answer annotations, associated lectures, and explanations.

\underline{GQA} \cite{hudson2019gqa}: A dataset specifically engineered for advanced real-world visual reasoning, utilizing scene graph-based structures to generate 22 million diverse, semantically-programmed questions. It incorporates novel evaluation metrics focusing on consistency, grounding, and plausibility, thereby establishing a rigorous standard for vision-language task assessment.

\underline{VisWiz} \cite{gurari2018vizwiz}: A visual question answering (VQA) dataset derived from naturalistic settings, featuring over 31,000 visual questions. It is distinguished by its goal-oriented approach, with images captured by blind individuals and accompanied by their spoken queries, along with crowdsourced answers.

\underline{CHAIR} \cite{rohrbach2018object}: A well-established benchmark for evaluating object hallucination in image captioning tasks, with two variants: CHAIR\textsubscript{I} and CHAIR\textsubscript{S}, which assess hallucination at the instance and sentence levels, respectively. we randomly sampled 500 images from the COCO \cite{lin2014microsoft} validation set and evaluated object hallucination using the CHAIR metric. Note that a lower CHAIR score indicates fewer hallucinations, which implies better alignment between the captions and the actual content of the images.
\begin{equation*}
\text{CHAIR}_I = \frac{\text{Number of hallucinated objects}}{\text{Number of all mentioned objects}},
\end{equation*}
\begin{equation*}
\text{CHAIR}_S = \frac{\text{Number of captions with hallucinated objects}}{\text{Number of all captions}}.
\end{equation*}

\subsection{Detail of Experimental Setup}
\partitle{Image augmentation strategies} 
\label{sec:strategies_detail}
We implement three effective image-side augmentation strategies to generate diverse responses from our model. By applying these techniques to the original images, we produce multiple distinct responses which are then synthesized into a comprehensive final output. This approach enhances model robustness by introducing controlled variations in visual input while maintaining semantic consistency. The augmentation strategies include:

\begin{itemize}
    \item Crop($s_\text{min},s_\text{max}$): Crop the image from minimum scale to the maximum scale ($s_\text{min}=0.2,s_\text{max}=0.5$ in our paper).
    \item Diffusion-S (Strong): Applies gaussian noise with 500 diffusion steps, creating significant but controlled perturbation.
    \item Diffusion-W (Weak): Introduces gaussian noise with 200 diffusion steps, offering a more moderate level of visual distortion.
    \item Contrast: Enhances image contrast by a factor of 2, accentuating visual boundaries and feature differences.
    \item Gamma: Performs gamma correction at a value of 0.8, lightening dark regions in the image. (Note that gamma values above 1 make shadows darker, while values below 1 make dark regions lighter).  
\end{itemize}

\begin{figure*}[tp]
\centering
\includegraphics[width=\linewidth]{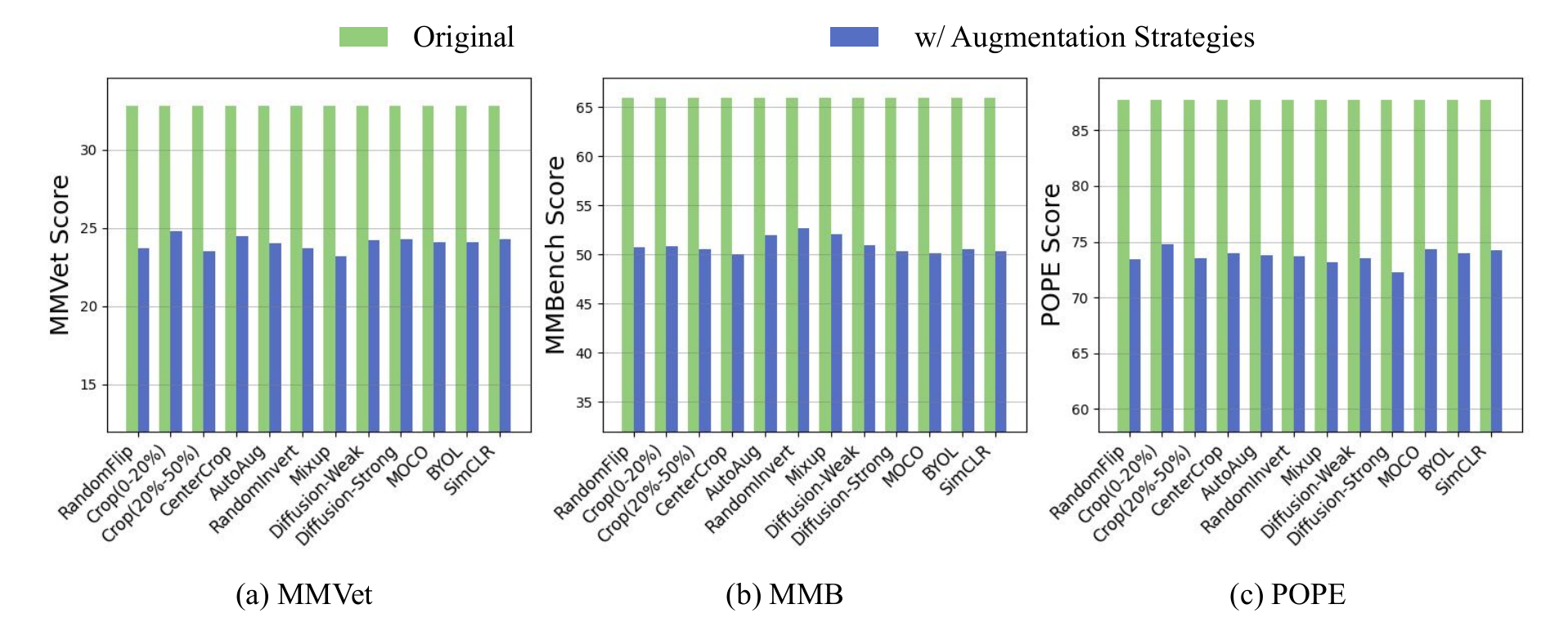}
    \caption{Comparison of 12 data augmentation strategies applied to LLaVA-1.5, including various geometric and color transformations as well as contrast learning enhancement methods. By analyzing these methods, the goal is to find the combination that best improves the performance of LVLMs.}
    \label{fig:strategies}
\end{figure*}

\partitle{Impact with Augmentation Strategies}
\label{sec:strategies_impact}
To better understand how different augmentation strategies influence performance, we conducted comprehensive analysis of 12 widely-used techniques from recent literature \citep{grill2020bootstrap, he2020momentum, chen2020simple, tong2024cambrian}, as shown in Figure \ref{fig:strategies}. Our investigation revealed important nuances in augmentation effectiveness. Not all augmentation methods provide equal benefits for model performance. Some strategies, particularly strong diffusion noise, proved excessively aggressive and interfered with the model's ability to extract and learn relevant visual features. Conversely, simpler augmentations like random flipping lacked sufficient complexity to drive meaningful performance improvements. Based on these findings, we carefully selected a combination of three augmentation strategies alongside the original images to strike the perfect balance.

\partitle{Evaluation of Scalability} 
To evaluate the scalability of \projectname~, we conducted experiments on the latest releases of DeepSeek-VL2-Tiny (based on DeepSeekMoE-3B) and DeepSeek-VL2 (based on DeepSeekMoE-27B). Table \ref{tbl:benchmark} presents quantitative improvements across all evaluation metrics, with more substantial gains observed on larger architectural configurations. When applied to DeepSeek-VL2, \projectname~ elevated MMVet scores from 56.9 to 60.2 and POPE scores from 92.5 to 96.2, indicating that the methodology accommodates scaling to larger architectures and addresses more complex analytical tasks.

\begin{table}[!h]
\centering
\vspace{-1.0em}
    \resizebox{0.48\textwidth}{!}{
    \large
    \begin{tabular}{lcccc}
        \toprule
       \textbf{Model} & \textbf{MMVet} &  \textbf{MMBench} & \textbf{MME} &  \textbf{POPE} \\
        \midrule
        DeepSeek-VL2-Tiny & 52.8  & 69.2 & 1,915 & 88.8 \\
        \quad + SeVa & 53.1 & 70.7 & 2,086 & 89.4  \\
        \quad + SHAPE & 55.1 & 73.5 & 2,103 & 94.7 \\
        \midrule
        DeepSeek-VL2 & 56.9 & 79.6 & 2,253 & 92.5  \\
        \quad + SeVa & 58.5 & 80.5 & 2,296 & 93.8  \\
        \quad + SHAPE & 60.2 & 81.0 & 2,318 & 96.2 \\
        \bottomrule
    \end{tabular}
    }
    \vspace{-1em}
\caption{Comparison of SHAPE with sota enhancement strategy applied to the DeepSeek.}
\label{tbl:benchmark}
\end{table}

\subsection{Exhibition Board}
\label{shape_case}

\begin{figure*}[tp]
\centering
\includegraphics[width=\linewidth]{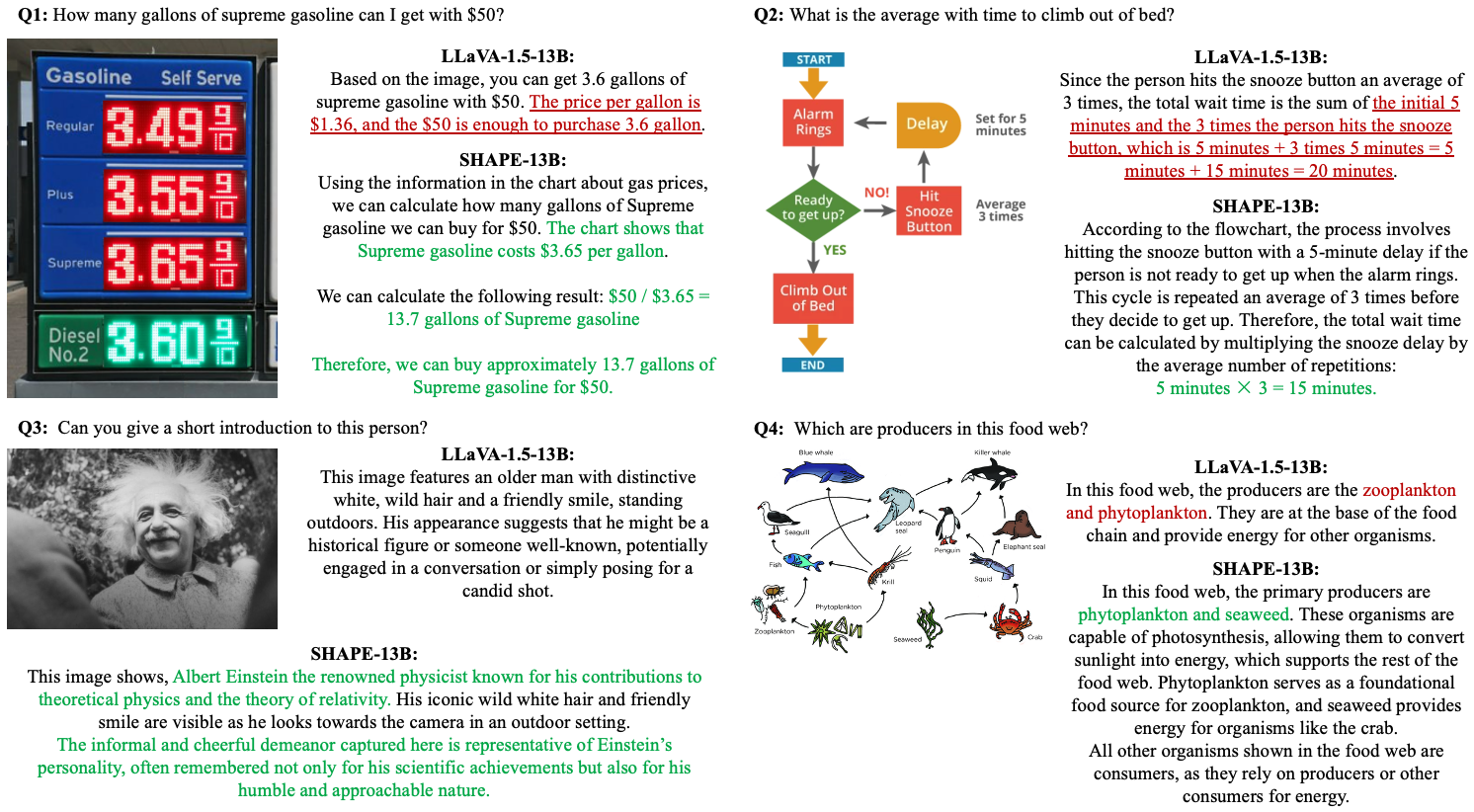}
    \caption{On multiple dimensions of the MMVET task (Recognition, OCR, Spatial Understanding, Mathematical Reasoning, Knowledge Retrieval and Generation), \projectname~ demonstrated significantly better results than LLaVA-1.5. To highlight these advancements, we provide direct comparisons with LLaVA-1.5’s original outputs. Incorrect responses from LLaVA-1.5 are marked in \textcolor{darkred}{red}, while our model’s more holistic answers are highlighted in \textcolor{darkgreen}{green} text.}
\label{fig:case}
\end{figure*}

This section presents \projectname~ is able to process the information in images more accurately, and provide contextually relevant answers in combination with a knowledge base, significantly improving the model's performance in complex visual-linguistic tasks, which can be seen in Figure \ref{fig:case}. This improvement reflects \projectname~'s higher semantic accuracy and stronger reasoning ability in image understanding and text generation tasks.

\subsection{GPT-4V-assisted Preference Annotations}
\label{sec:gptjudge}

\begin{figure*}[tp]
    \centering
    \includegraphics[width=\linewidth]{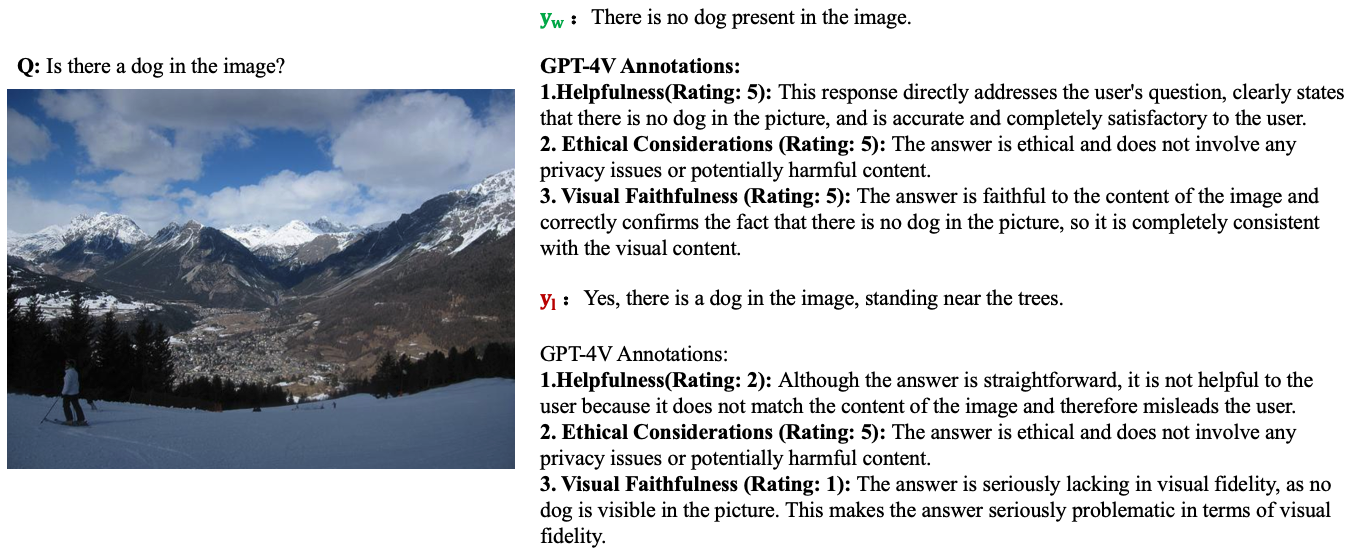}
    \caption{Leveraging GPT-4V for Accurate Visual Content Evaluation: Confirming the Presence or Absence of Objects.}
    \label{fig:gpt1}
\end{figure*}

\begin{figure*}[tp]
    \centering
    \includegraphics[width=\linewidth]{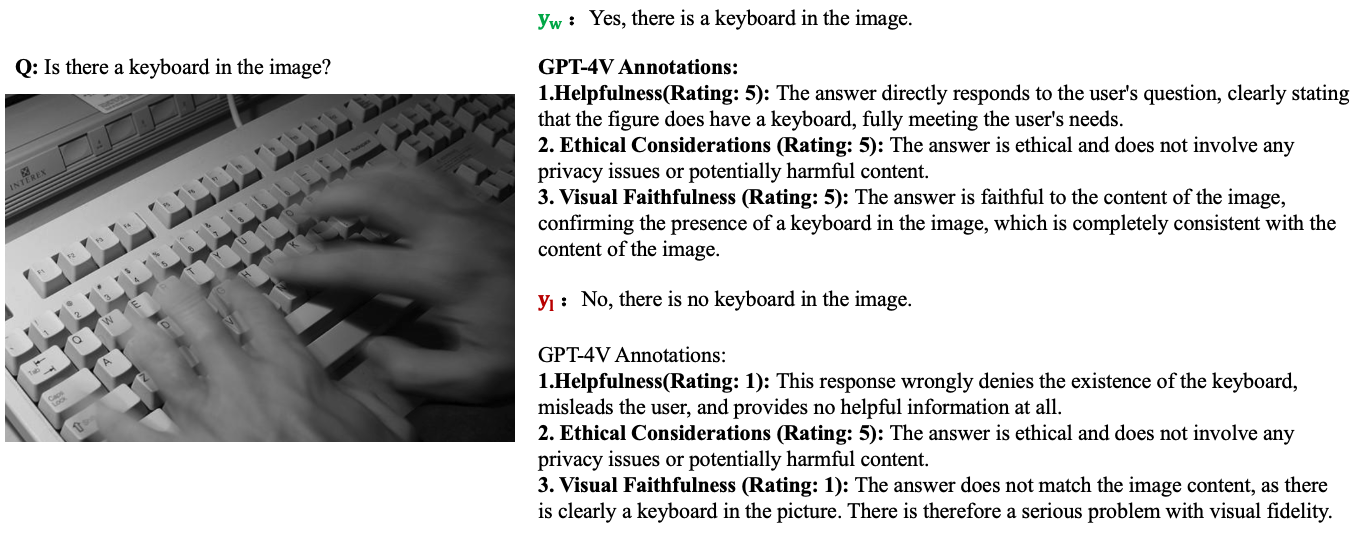}
    \caption{Ensuring Visual Faithfulness and Helpfulness in Image-Text Alignment Using GPT-4V Scoring.}
    \label{fig:gpt2}
\end{figure*}

GPT-4V's role as a judge is widely recognized as reliable due to its superior capabilities in visual and linguistic comprehension, enabling a thorough assessment of different models' performance in multimodal tasks \citep{bai2022training,lee2023rlaif,cui2023ultrafeedback}. We present two examples the GPT-4V annotations in the Figure \ref{fig:gpt1} and \ref{fig:gpt2}.

\end{document}